\documentclass[english]{article}
\usepackage[T1]{fontenc}
\usepackage[latin9]{inputenc}
\usepackage{float}
\usepackage{hyperref}
\usepackage{amsmath}
\usepackage{amssymb}
\usepackage{graphicx}
\usepackage{babel}

\title{Sparse 3D convolutional neural networks}
\author{Ben Graham\\
{\small{University of Warwick}}\\
\texttt{\small{b.graham@warwick.ac.uk}}}

\begin{document}

\maketitle
\begin{abstract}
We have implemented a convolutional neural network designed for processing
sparse three-dimensional input data. The world we live in is three
dimensional so there are a large number of potential applications including 3D object recognition and analysis of space-time objects.
In the quest for efficiency, we experiment with CNNs on the 2D triangular-lattice
and 3D tetrahedral-lattice.
\end{abstract}

\section{Convolutional neural networks}

Convolutional neural networks (CNNs) are powerful tools for understanding
data with spatial structure such as photos. They are most commonly
used in two dimensions, but they can also be applied more generally.
One-dimensional CNNs are used for processing time-series such as human
speech \cite{lecun-bengio-95a}. Three dimensional CNNs have been
used to analyze movement in 2+1 dimensional space-time\cite{3dConvolutions,KarpathyCVPR14_LargeScaleVideoClassificationWithCNNs}
and for helping drones find a safe place to land \cite{maturana_icra_2014}.
Three dimensional convolutional deep belief networks have been used to recognize objects in 2.5D depth maps \cite{ShapeNet_Wu_2015_CVPR}.

In \cite{GrahamSparse}, a \emph{sparse} two-dimensional CNN is implemented
to perform Chinese handwriting recognition. When a handwritten character
is rendered at moderately high resolution on a two dimensional grid,
it looks like a sparse matrix. If we only calculate the hidden units
of the CNN that can actually \emph{see} some part of the input field
the pen has visited, the workload decreases. We have extended this
idea to implement sparse 3D CNNs %
\footnote{Software for creating sparse 2, 3 and 4 dimensional CNNs is available
at \url{https://github.com/btgraham/SparseConvNet}}. Moving from two to three dimensions, the \emph{curse of dimensionality}
becomes relevant---an $N\times N\times N$ cubic grid contains many
more points than an $N\times N$ square grid. However, the curse can
also be taken to mean that the higher the dimension, the more likely
interesting input data is to be sparse.

To motivate the idea of a sparse 3D CNN, imagine you have a loop of
string with a knot in it. Mathematically, detecting and classifying
knots is a hard problem; a piece of string can be very tangled without
actually being knotted. Suppose you are only interested in
`typical' knots---humans can quite easily learn to spot the difference
between, say, a trefoil knot and a figure of eight knot. If you want
to take humans out of the loop, then you could train a 2D CNN to recognize and classify pictures of knots.
However, pictures taken from certain angles will not contain enough
information to classify the knot due to parts of the string being
obscured. Suppose instead that you can trace the path of the string
through three dimensional space; you could then use a 3D CNN to classify the knot. The string is essentially one dimensional, so the parts of space that the string visits will be sparse.

The example of the string is just a thought experiment. However, there
are many real-world problems, in domains such as robotics and biochemistry,
where understanding 3D structure is important and where sparsity is
applicable.

\begin{figure}
\begin{centering}
\includegraphics[width=0.3\columnwidth]{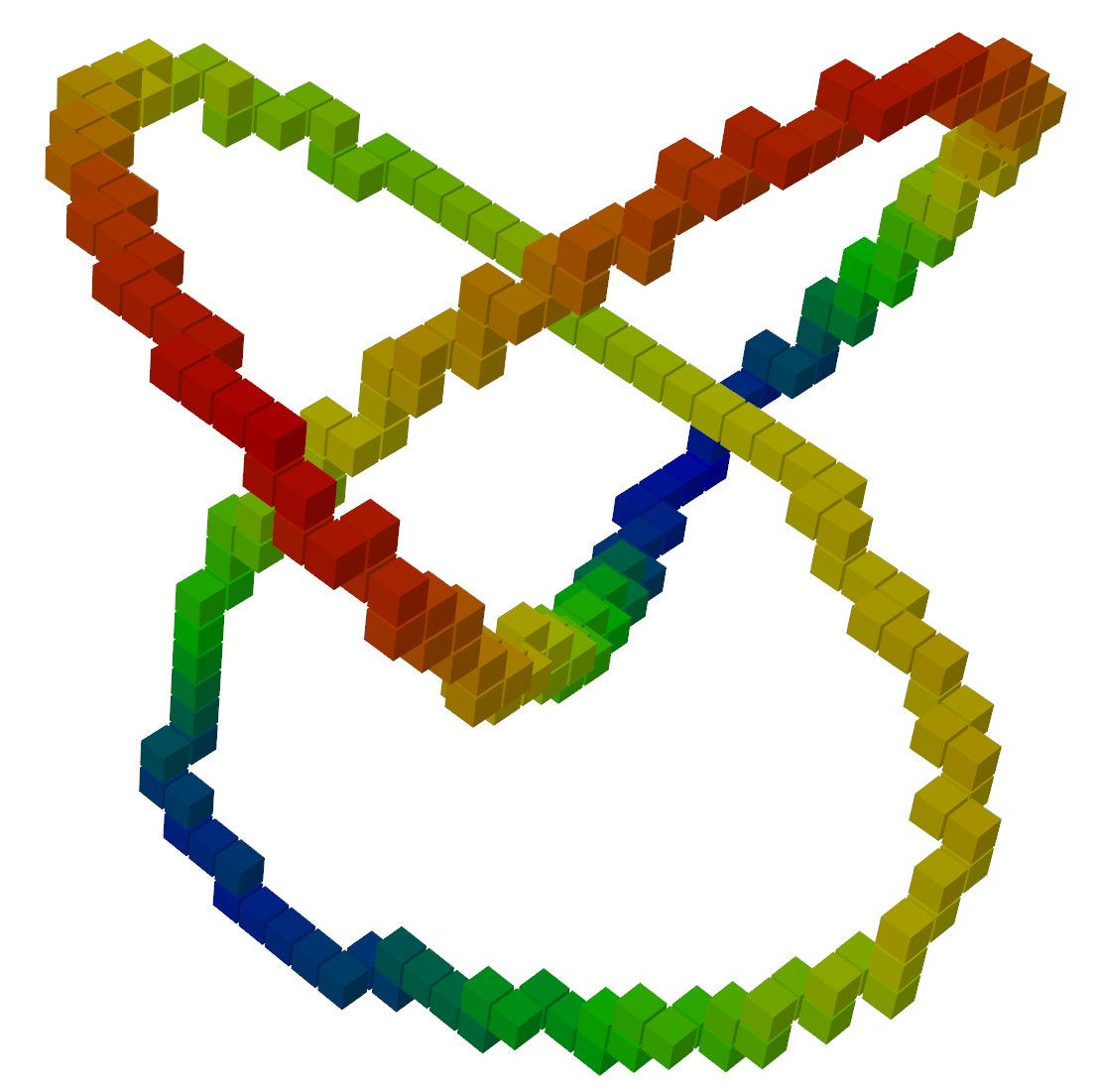}\includegraphics[width=0.3\columnwidth]{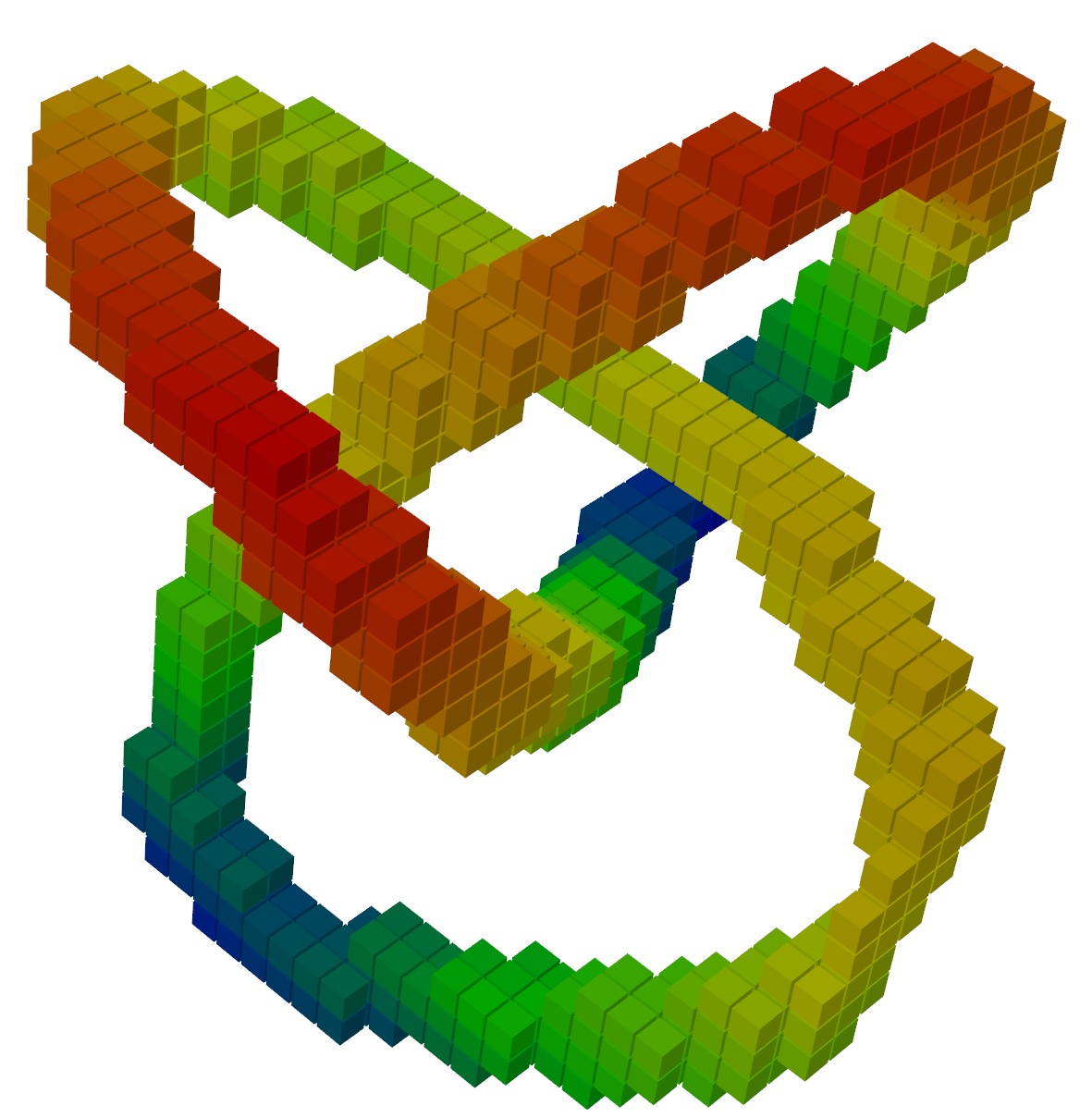}\includegraphics[width=0.3\columnwidth]{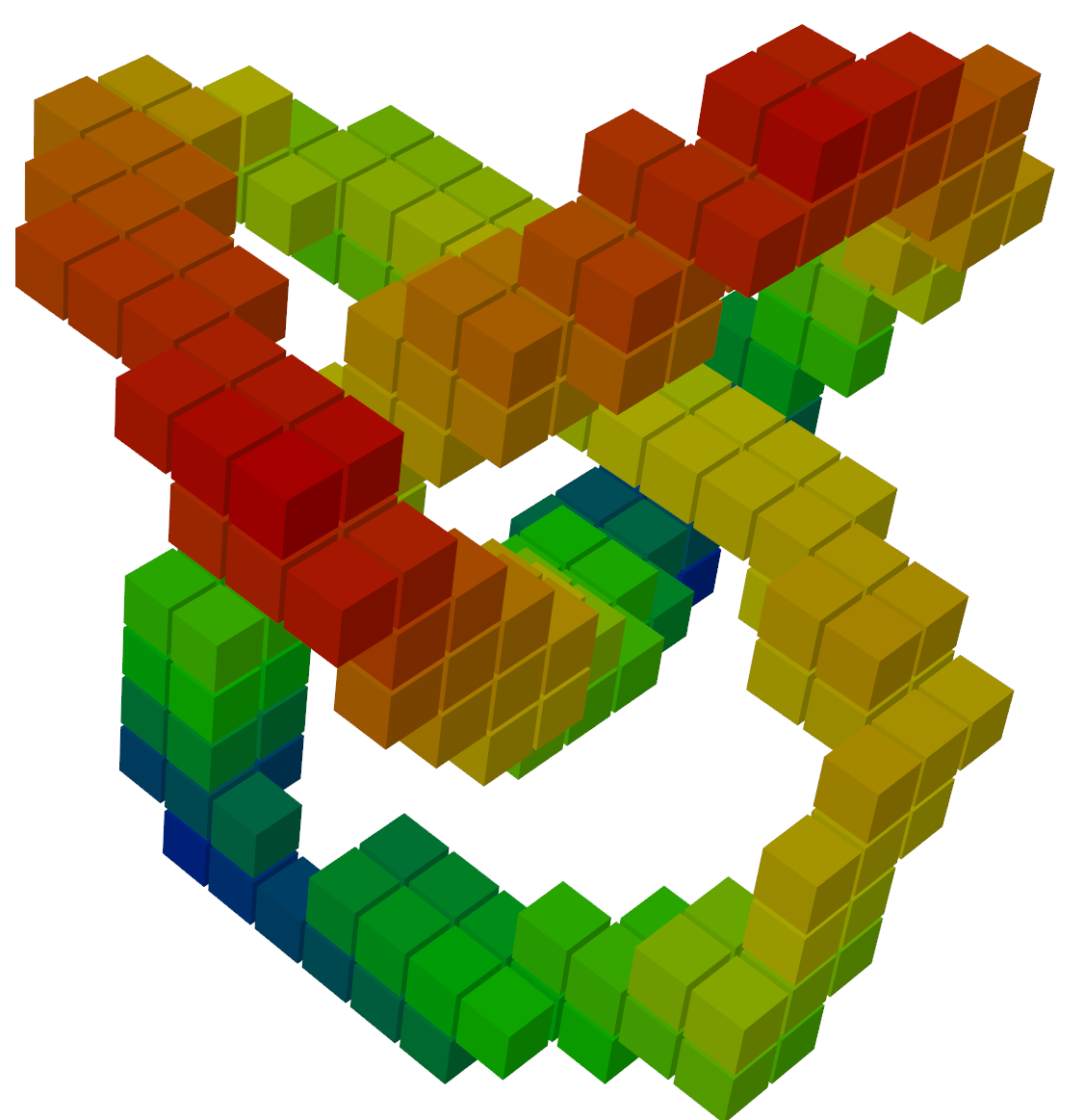}
\par\end{centering}

\caption{Left to right: A trefoil knot has been drawn in the cubic lattice;
these are the input layer's active sites. Applying a $2\times2\times2$
convolution, the number of active (i.e. non-zero) sites increases.
Applying a $2\times2\times2$ pooling operation reduces the scale, which tends to decrease the number
of active sites.\label{fig:Trefoil}}
\end{figure}

\subsection{Adding a dimension to 2D CNNs?}

Recently there has been an explosion of research into conventional
two-dimensional CNNs. This has gone hand-in-hand with a substantial increase
in available computing power thanks to GPU computing. For photographs
of size $224\times224$, evaluating model C of \cite{DelvingDeeperRectifiers}'s
19 convolutional layers requires 53 billion multiply-accumulate operations.

Although model C's input is represented as a 3D array of size $224\times224\times3$,
it is still fundamentally 2D---we can think of it as a 2D array of
vectors, with each vector storing an RGB-color
value. Model C's initial convolutional layer consists of 96 convolutional
filters of size $7\times7$, applied with stride 2. Each filter is
therefore applied $(224/2)^{2}$ times.

This makes 3D CNNs sound like a terrible idea. Consider adapting model
C's network architecture to accept 3D input with size $224\times224\times224\times3$,
i.e. some kind of 3D model where each points has a color. To apply a $7\times7\times7$
convolutional filter with the same stride, we would need to apply
it 112 more times than in the 2D case, with each application requiring
7 times as many operations. Extending the whole of model C to 3D would
increase the computational complexity to 6.1 trillion operations.
Clearly if we want to use 3D CNNs, then we need to do some things
differently.

Applying the convolutions in Fourier space \cite{LecunFastTrainingFFTConvolutions}, or using separable filters \cite{rigamonti2013learning} could help, but simply the amount of memory needed to store large 3D grids of vectors would still be a problem.
Instead we try two things that work well together.
Firstly we use much smaller filters, using network architectures similar to the ones introduced in \cite{multicolumndeep}.
The smallest non-trivial filter possible on a cubic lattice has size $2\times2\times2$, covering $2^{3}=8$ input sites.
In an attempt to improve efficiency, we will also consider the tetrahedral lattice,
where the smallest filter is a tetrahedron of size 2 which covers
just 4 input sites. Secondly, we will only consider problems where the
input is sparse. This saves us from having to have the convolutional filters
visit each spatial location. If the interesting part of the input is a 1D curve or a 2D surface,
then the majority of the 3D input field will receive only zero-vectors for inputs. Sparse CNNs are more efficient when used with smaller filters, as the hidden layers tend to be sparser.

\subsection{CNNs on different lattices}

\begin{figure}
\begin{centering}
\includegraphics[width=1\textwidth]{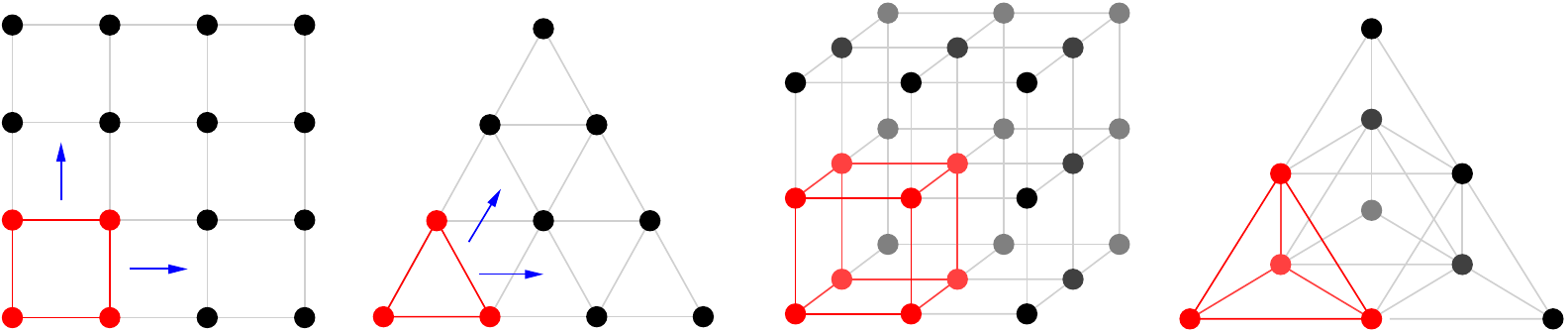}
\par\end{centering}

\begin{centering}
(i)$\hspace{1em}\hspace{1em}\hspace{1em}\hspace{1em}\hspace{1em}\hspace{1em}\hspace{1em}$
(ii)$\hspace{1em}\hspace{1em}\hspace{1em}\hspace{1em}\hspace{1em}\hspace{1em}\hspace{1em}$
(iii)$\hspace{1em}\hspace{1em}\hspace{1em}\hspace{1em}\hspace{1em}\hspace{1em}\hspace{1em}$
(iv)
\par\end{centering}

\caption{Convolutional filter shapes for different lattices: (i) A $4\times4$
square grid with a $2\times2$ convolutional filter. (ii) A triangular
grid with size 4, and a triangular filter with size 2. (iii) A $3\times3\times3$
cubic grid, and a $2\times2\times2$ filter. (iv) A tetrahedral grid
with size 3, and a filter of size 2.\label{fig:Convolutional-filter-shapes}}
\end{figure}

Each layer of a CNN consists of a finite graph, with a vector of input/hidden
units at each site. For regular two dimensional CNNs, the graphs are
square grids. The convolutional filters are square-shaped too, and
they move over the underlying graph with two degrees of freedom; see
Figure \ref{fig:Convolutional-filter-shapes} (i). Similarly, 3D CNNs
are normally defined on cubic grids. The convolutional filters are
cube-shaped, and they move with three degrees of freedom; see Figure
\ref{fig:Convolutional-filter-shapes} (iii).

In principle we could also build 4D CNNs on hypercubic grids, and
so on. However, as the dimension $d=2,3,4,...$ increases, the size
$2^{d}$ of the smallest non-trivial filter is growing exponentially.
In the interests of efficiency, we will also consider CNNs with a
different family of underlying graphs. In 2D, we can build CNNs based
on triangles. For each layer, the underlying graph is a triangular
grid, and the convolutional filters are triangular, moving with two
degree of freedom; see Figure \ref{fig:Sparse-convolution.-On} (ii).
In 3D, we can use a tetrahedral grid and tetrahedral filters that
move with three degrees of freedom; see Figure \ref{fig:Sparse-convolution.-On}
(iv). We could extend this to 4D with hypertetrahedrons, etc. In $d$
dimensions, the smallest convolutional filters contain only $d+1$
sites, rather than exponentially many.

To describe CNN architecture on these different lattices, we will
still use the common ``$n$C$f/s$-MP$p/s$-...'' notation. The
$n$ counts the number of convolutional filters, $f$ measures the
linear size of the filters---the number of input sites the convolutional
filter covers is $f^{2}$, $f^{3}$, $\binom{f+1}{2}$, $\binom{f+2}{3}$
on the square, cubic, triangular and tetrahedral lattices, respectively---and
$s$ denotes the stride. The $p$ measures the linear size of the
max-pooling regions. The $/s$ is omitted when $s=1$ for convolutions
or $s=p$ for pooling. For example, on the tetrahedral lattice $32\mathrm{C}2-\mathrm{MP}3/2$
means 32 filters of size 2 which cover $\binom{2+2}{3}=4$ input sites,
followed by max pooling with pooling regions of size $\binom{3+2}{3}=10$,
and with adjacent pooling regions overlapping by one.

\subsection{Sparse operations}

Sparse CNNs can be thought of as an extension of the idea of sparse
matrices. If a large matrix only has small number of non-zero entries
per row and per column, then it makes sense to use a special data
structure to store the non-zero entries and their locations; this
can both dramatically reduce memory requirements and speed up operations
such as matrix multiplication. However, if 10\% of the entries are
non-zero, then the advantages of sparsity may be outweighed by the
efficiency which which dense matrix multiplication can be carried
out, either using Strassen's algorithm, or optimized GPU kernels.

\begin{figure}
\begin{centering}
\includegraphics[width=0.5\columnwidth]{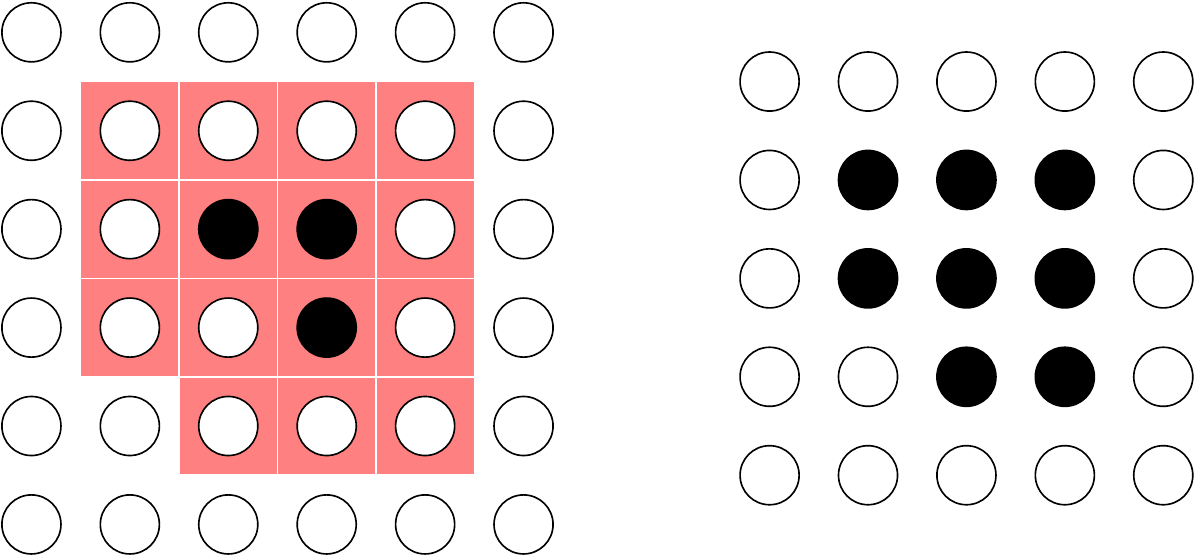}
\par\end{centering}

\caption{Calculating a $2\times2$ convolution for a sparse CNN: On the left
is a $6\times6$ square grid with 3 active sites. The convolutional
filter needs to be calculated at each location that covers at least
on active site; this corresponds to the shaded region. The figure
on the right marks the location of the eight active sites in the $5\times5$
output layer. Sparsity decreases with each convolution and pooling
operation. However, a CNN spends most of its time processing the lower
layers, so sparsity can still be useful.\label{fig:Sparse-convolution.-On}}
\end{figure}

The sparse CNN algorithm from \cite{GrahamSparse} can be tweaked
to work efficiently on general lattices.
The spatial size of each of the CNN's data layers is described by a lattice-type graph (similar to the ones in Figure ~\ref{fig:Convolutional-filter-shapes}).
At each spatial location
in the grid, there is a dimension-less vector of input or hidden units.
Depending on the input, some of the spatial locations will be defined
to be active.
\begin{itemize}
\item A spatial location in the input layer graph is declared \emph{active} if the location's vector is not the zero vector.
\item Declare that a spatial location in a hidden layer is active if \emph{any}
of the spatial location in the layer below from which it receives
input are active.
\end{itemize}
See Figure \ref{fig:Sparse-convolution.-On} for a 2D example of a
sparse convolution, and see Figure \ref{fig:Trefoil} for a 3D example.

By induction, the dimension-less vectors at each non-active spatial location in the \mbox{$n$-th} hidden layer are all the same; the shared value of the vectors can be pre-computed. We will call this the {\em ground state} vector for the $n$-th level. The ground state for the input layer is just the zero vector.

We will now describe the implementation of the sparse convolution for the types of graphs shown in Figure \ref{fig:Convolutional-filter-shapes}. For simplicity, we will focus on case (i), the 2D square grid; the other cases are very similar.

Suppose that an image has input field size $m_\mathrm{in}\times m_\mathrm{in}$,
and that the number of active spatial locations is $a_{\mathrm{in}}\in\{0,1,\dots,m_{\mathrm{in}}^{2}\}$.
Suppose an $f\times f$ convolutional filter will act on the image,
and let $n_{\mathrm{in}}$ and $n_{\mathrm{out}}$ denote the number
of input and output features per spatial location. The input to the
first convolutional operation consists of:
\begin{itemize}
\item A matrix \emph{$M_{\mathrm{in}}$} with size $a_{\mathrm{in}}\times n_{\mathrm{in}}$.
Each row corresponds to the vector at one of the active spatial locations.
\item A map or hash table $H_{\mathrm{in}}$ of (\emph{key},\emph{value}) pairs. The \emph{keys} are the active
spatial locations. The \emph{values} record the number of the corresponding
row in $M_{\mathrm{in}}$.
\item The input layer's ground state vector $g_\mathrm{in}$.
\item An $(f^{2}n_{\mathrm{in}})\times n_{\mathrm{out}}$ matrix \emph{$W$}
containing the weights that define the convolution.
\item A vector $B$ of length $n_{\mathrm{out}}$ specifying the values of the bias units.
\end{itemize}
To calculate the output of the first hidden layer:
\begin{enumerate}
\item Iterate through $H_\mathrm{in}$ and determine the number $a_{\mathrm{out}}$
of active spatial locations in the output layer. A site in the output
layer is active if any of the input sites are active. Build a hash
table $H_{\mathrm{out}}$ to uniquely identify each of the active
output spatial locations with one of the integers $1,2,\dots a_{\mathrm{out}}$.
\item Use $H_{\mathrm{in}}$, $M_{\mathrm{in}}$ and $g_\mathrm{in}$ to build a matrix $Q$
of size $a_\mathrm{out}\times(f^{2}n_{\mathrm{in}})$; each row of $Q$ should
correspond to the inputs visible to the convolutional filter at the
corresponding output spatial location.
\item Calculate $M_{\mathrm{out}}=Q\times W+B$.
\end{enumerate}
We implemented step 1 on the CPU and steps 2 and 3 on the GPU.
If $W$ is small, the computational bottleneck will be I/O-related, steps 1 and 2.
If $W$ is large, the bottleneck will be performing the dense matrix
multiplication in step 3 to calculate $M_{\mathrm{out}}$.

The procedure for max-pooling is similar. Max-pooling is always I/O-bound.

\section{Experiments}

We have performed experiments to test triangular and sparse 3D CNNs.
Unlike the 2D case, there are not yet any standard benchmarks for
evaluating 3D CNNs, so we just picked a range of different types of
data. When faced with a trade-off between computational cost and accuracy,
we have preferred to train smaller network to see what can be achieved
on a limited computational budget, rather than trying to maximize
performance at any cost.

For some of the experiments we used $n$-fold repetitive testing:
we processed each test case $n$ times, with some form of data augmentation,
for $n$ a small integer, and averaged the output.

\subsection{Square versus triangular 2D convolutions}

As a sanity test regarding our unusually shaped CNNs, we first did a 2D
experiment with the CIFAR-10 dataset of small pictures \cite{CIFAR10}
to compare CNNs on the square and triangular lattices. We will call
the networks SquareNet and TriangLeNet, respectively. Both networks
have 12 small convolutional layers split into pairs by 5 layers of
max-pooling, and with the $n$-th pair of convolutional filters each
having 32$n$ output features:
\[
32C2-32C2-MP3/2-\dots-MP3/2-192C2-192C2-\mathrm{output}
\]
We extended the training data using affine transformations. For the
triangular lattice, we converted the images to triangular coordinates
using an additional affine transformation. See Table \ref{tab:Comparison-between-square}
for the results.

TriangLeNet has a computational cost that is 26\% lower than the more
conventional SquareNet. In terms of test errors, there does not seem
to be any real difference between the two networks.

\begin{table}
\centering{}%
\begin{tabular}{|cccc|}
\hline
CNN & MegaOps & test error & 12-fold test error\tabularnewline
\hline
SquareNet & 41 & 9.24\% & 7.66\%\tabularnewline
TriangLeNet & 30 & 9.70\% & 7.50\%\tabularnewline
\hline
\end{tabular}\caption{Comparison between square and triangular 2D CNNs for CIFAR-10\label{tab:Comparison-between-square}}
\end{table}

\subsection{Object recognition}\label{sec:shrec}

\begin{figure}
\begin{centering}
\includegraphics[angle=90,width=0.2\textwidth]{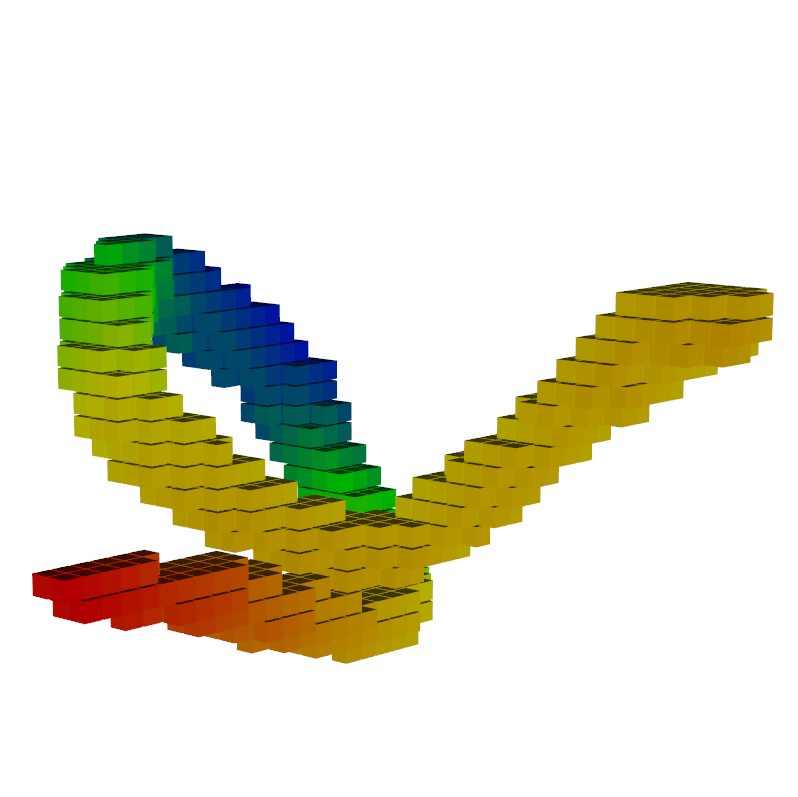}\includegraphics[width=0.2\textwidth]{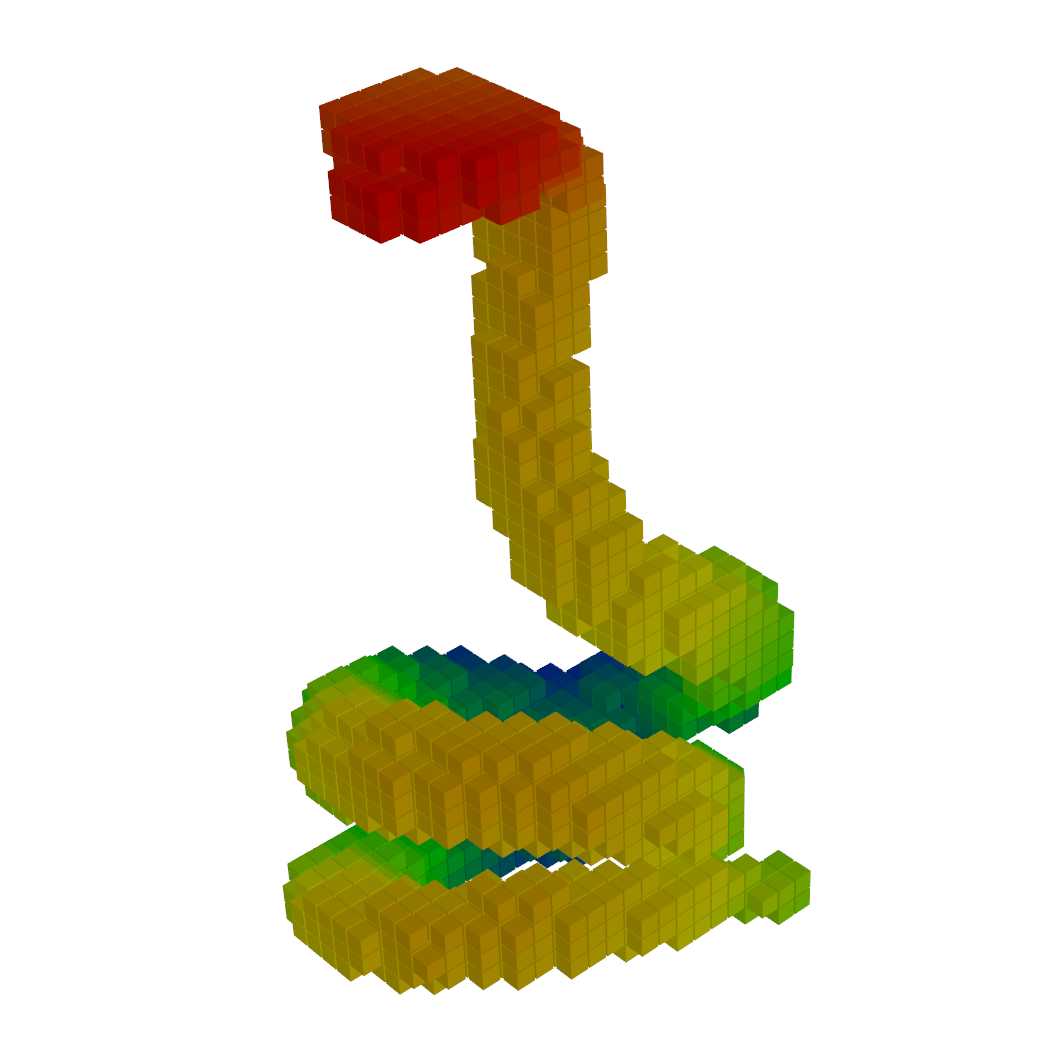}\includegraphics[width=0.2\textwidth]{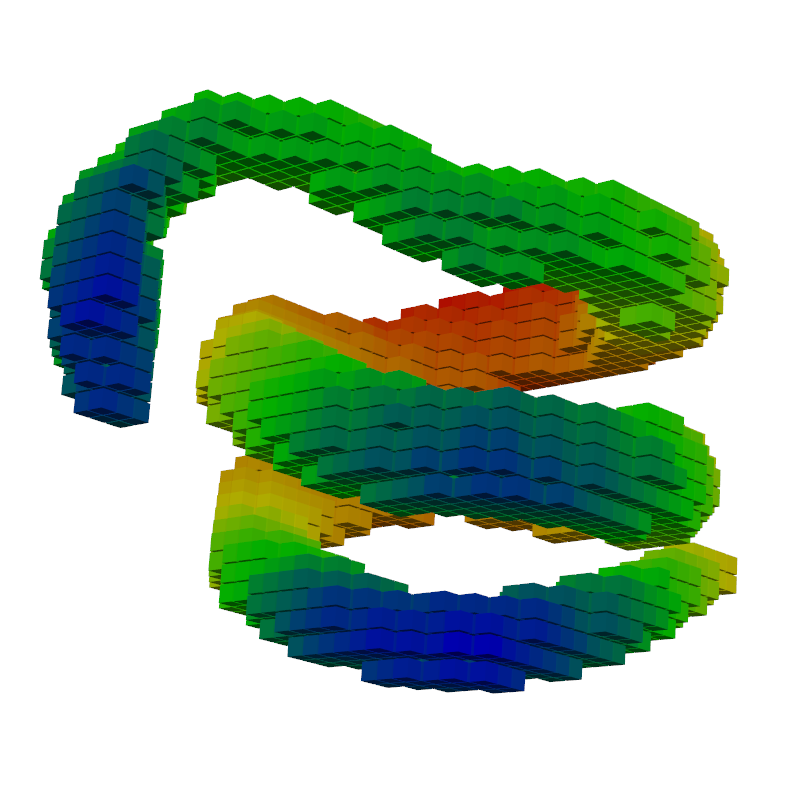}\includegraphics[width=0.2\textwidth]{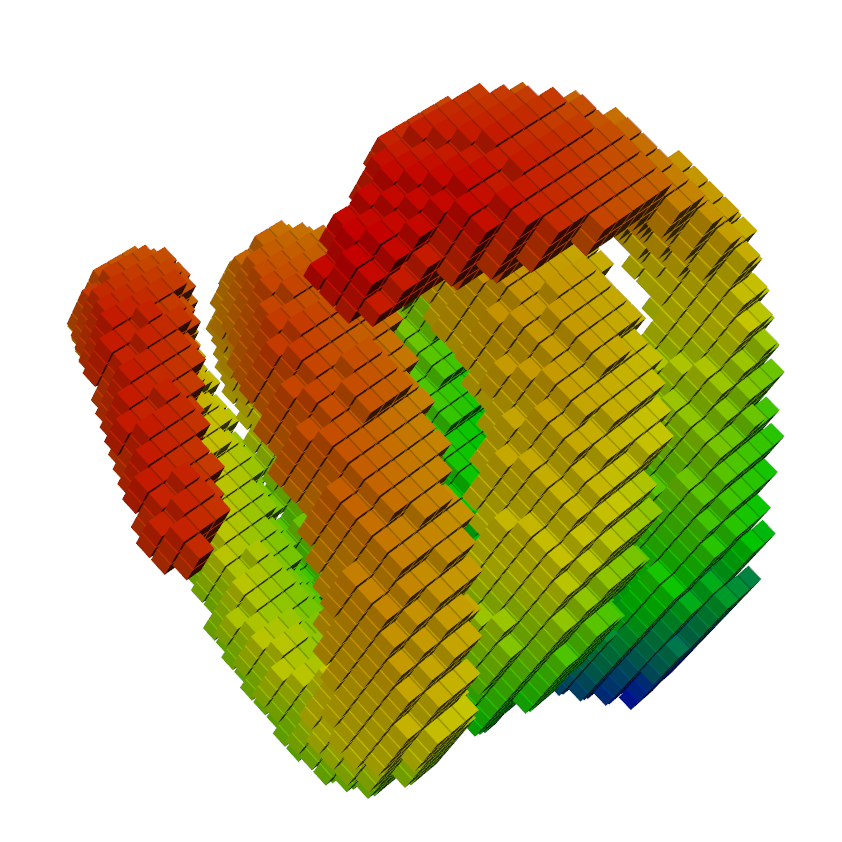}
\par\end{centering}

\begin{centering}
\includegraphics[width=0.2\textwidth]{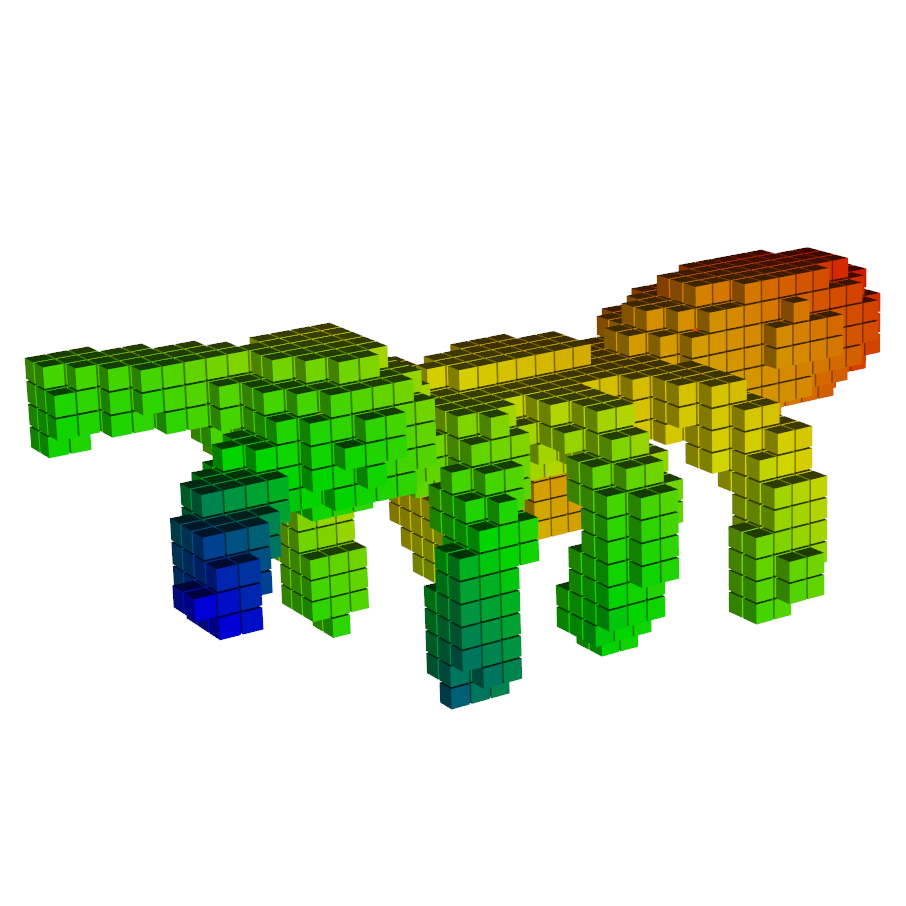}\includegraphics[width=0.2\textwidth]{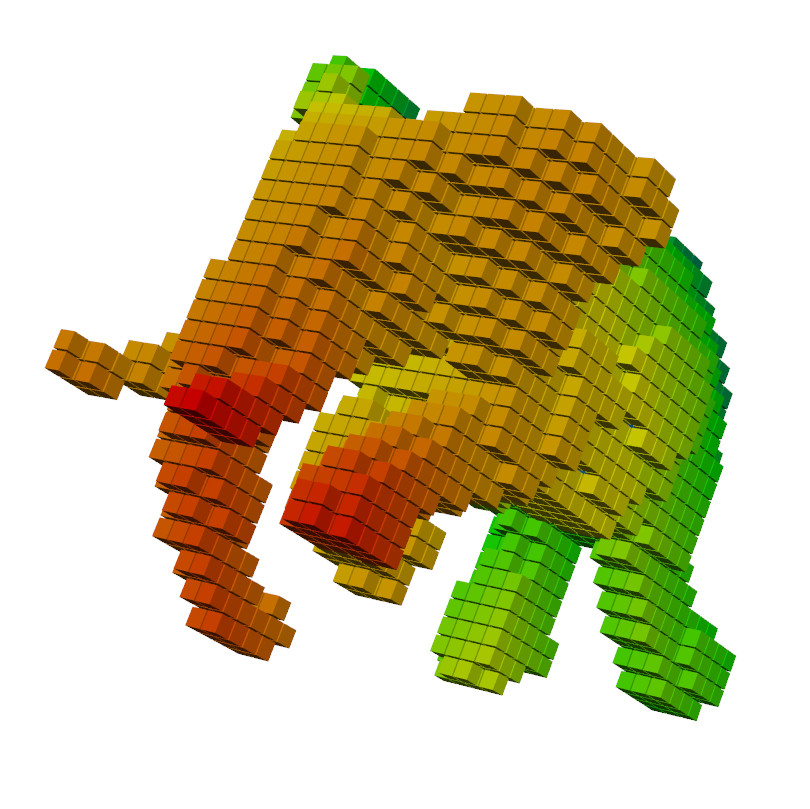}\includegraphics[width=0.2\textwidth]{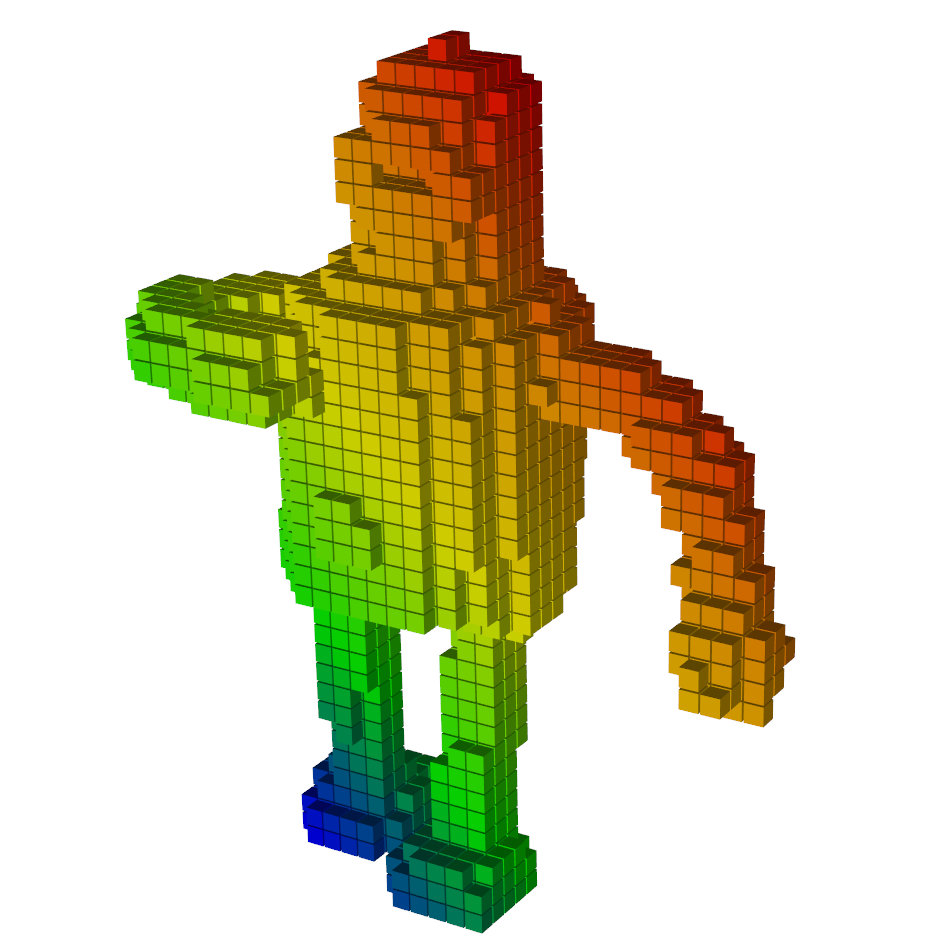}\includegraphics[width=0.2\textwidth]{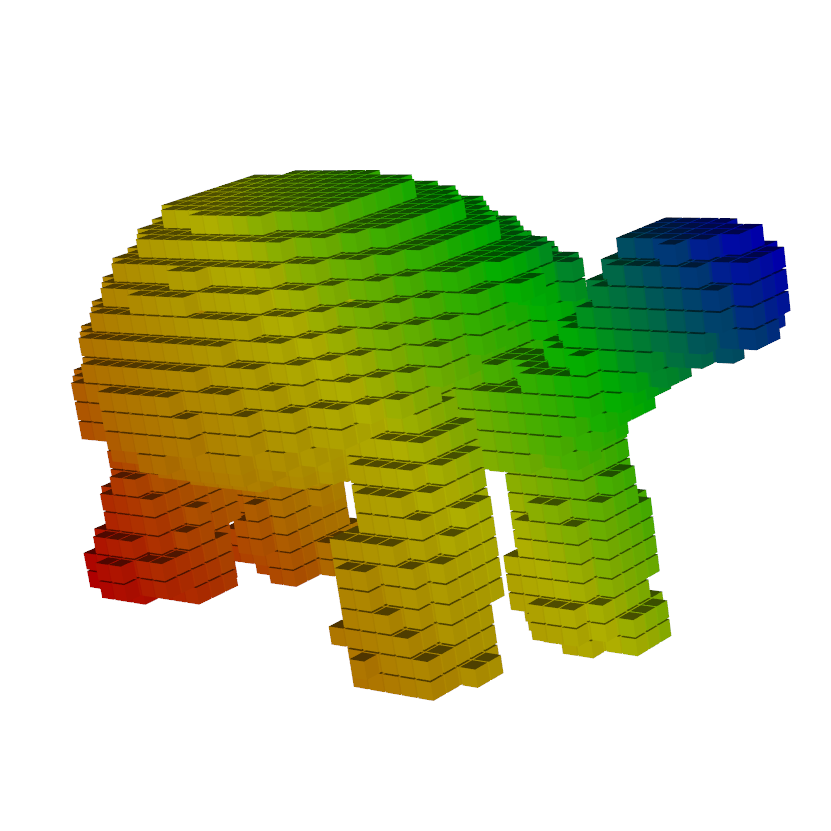}
\par\end{centering}

\caption{Items from the 3D object dataset used in Section \ref{sec:shrec}, embedded into a
 $40\times40\times40$ cubic grid. Top row: four items from the snake class.
Bottom row: an ant, an elephant, a robot and a tortoise. \label{fig:shrec} }
\end{figure}

To test the 3D CNN, we used a dataset of  3D objects\footnote{SHREC2015 Non-rigid 3D Shape Retrieval dataset \url{http://www.icst.pku.edu.cn/zlian/shrec15-non-rigid/data.html}}, each stored as a mesh of triangles in the OFF-file format.
The dataset contains 1200 exemplars split evenly between 50 classes
(aliens, ants, armadillo, ...). The dataset was intended to be used
for unsupervised learning, but as CNNs are most often used for supervised
learning, we used 6-fold cross-validation to measure the ability of
our 3D CNNs to learn shapes. To stop the dataset being too easy, we
randomly rotated the objects during training and testing. This is to force the CNN to truly learn to recognize shape, and not rely on some classes of objects tending to have a certain orientation.

\begin{figure}
\noindent \centering{}%
\begin{minipage}[t][1\totalheight][c]{0.45\columnwidth}%
\vspace{2mm}
\noindent \begin{center}
\resizebox{\textwidth}{!}{%
\begin{tabular}{|cccc|}
\hline
scale & pooling & $\times10^{6}$ operations & tests/s\tabularnewline
\hline
20 & $4\times$MP3/2$\triangle$ & 9 & 1133\tabularnewline
40 & $5\times$MP3/2$\triangle$ & 35 & 675\tabularnewline
80 & $6\times$MP3/2$\triangle$ & 143 & 286\tabularnewline
20 & $4\times$MP3/2 $\square$ & 36 & 1190\tabularnewline
40 & $5\times$MP3/2 $\square$ & 126 & 794\tabularnewline
80 & $6\times$MP3/2 $\square$ & 406 & 310\tabularnewline
20 & $6\times$FMP$\square$ & 116 & 1100\tabularnewline
32 & $7\times$FMP$\square$ & 279 & 849\tabularnewline
\hline
\end{tabular}}
\par\end{center}%
\end{minipage}~~~~%
\begin{minipage}[t][1\totalheight][c]{0.55\columnwidth}%
\noindent \begin{center}
\includegraphics[bb=0bp 0bp 432bp 280bp,width=1\textwidth]{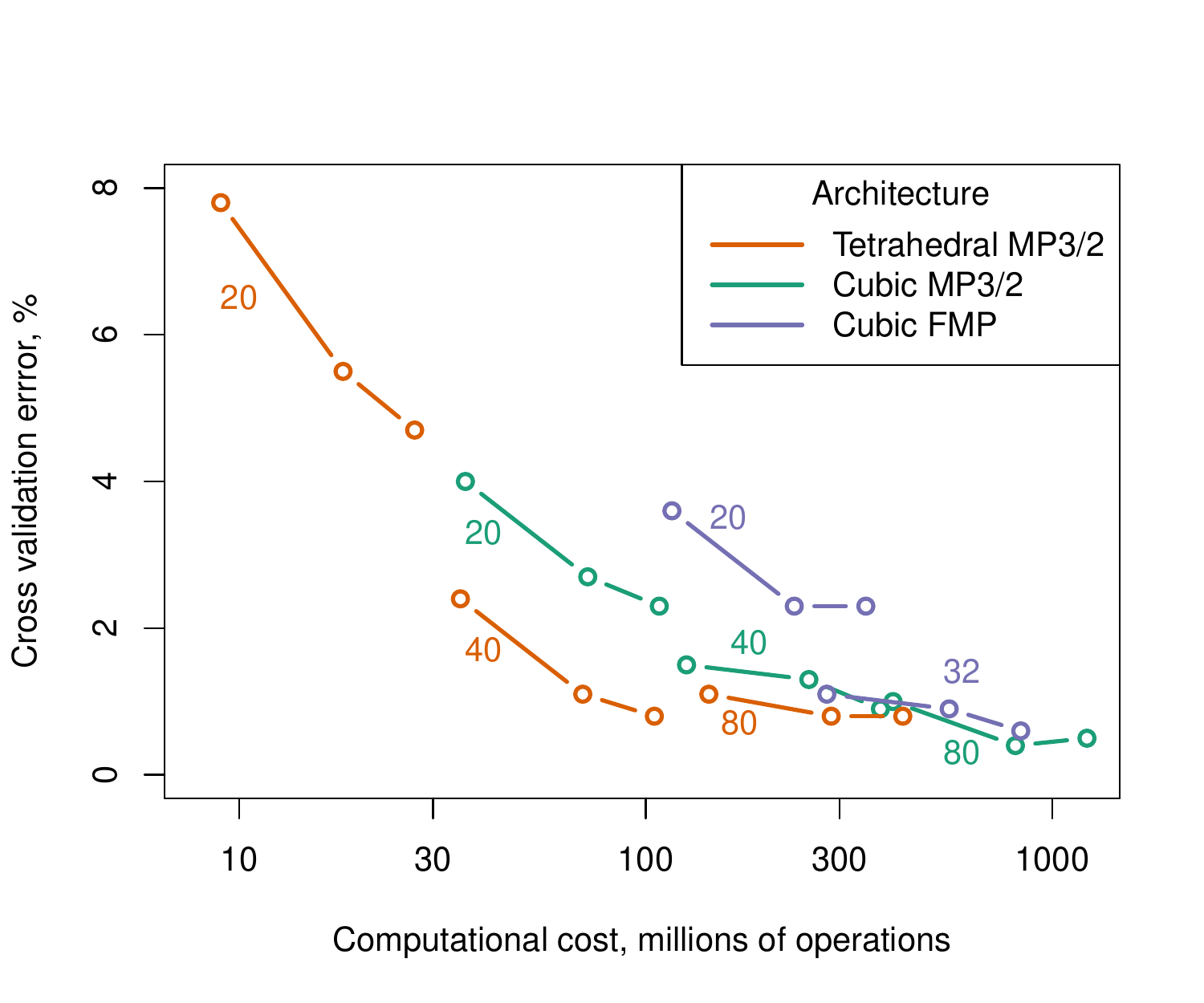}
\par\end{center}%
\end{minipage}\caption{6-fold cross-validation error rate for 3D object recognition for different
CNN architectures. The lines in the graph correspond to performing
1-, 2- and 3-fold testing with a given CNN. The table given the computational
complexity and speed of the network on a Nvidia GeForce GTX 780 GPU.
\label{fig:SHREC2015-supervised-learning.}}
\end{figure}

All the CNNs we tested took the form
\[
32\mathrm{C}2-\mathrm{pooling}-64\mathrm{C}2-\mathrm{pooling}-96\mathrm{C}2-...-\mathrm{output.}
\]
We rendered the 3D models at a variety of different scales, and varied
the number of levels of pooling accordingly. We tried using MP3/2
pooling on the cubic and tetrahedral lattices. We also tried a stochastic
form of max-pooling on the cubic lattice which we denote FMP\cite{FractionalMaxPooling}; we used FMP to downsample the hidden layer by a factor of $2^{2/3}\approx1.59$; this allows us to gently increase
the number of learnt layers for a given input scale. See Figure \ref{fig:SHREC2015-supervised-learning.}.

The tetrahedral CNNs are substantially cheaper computationally, but
less accurate at the smallest scale. The FMP pooling provides the
highest accuracy when the scale is small, but they are quite a bit
more expensive. If we look at the number of test samples that can
be processed per second, we see that for such small CNNs the calculations
are actually I/O-bound, so tetrahedral network is not as much faster
as we might have expected based on the computational cost. However,
with a less powerful processor, it is likely that there would be a
speed advantage to the tetrahedral lattice.

\subsection{2D space + 1D time = 3D space-time}

The CASIA-OLHWDB1.1 database contains online handwriting samples of
the 3755 GBK level-1 Chinese characters \cite{CASIA}. There are approximately
240 training characters, and 60 test characters, per class. \emph{Online}
means that the pen strokes were recorded in the order they were made.

A test error of 5.61\% is achieved by drawing the characters with
size $40\times40$ and learning to recognize their pictures with a
2D CNN \cite{multicolumndeep}. Evaluating that network's four convolutional
layers requires 72 million multiply-accumulate operations.

With a 3D CNN, we can use the order in which the strokes were written
to represent each character as a collection of paths in 2+1 dimensional
space-time with size $40\times40\times40$. A 3D CNN with architecture
\[
32\mathrm{C}3-\mathrm{MP}3/2-64\mathrm{C}2-\mathrm{MP}3/2-128\mathrm{C}2-\mathrm{MP}3/2-256\mathrm{C}2-\mathrm{MP}3/2-512\mathrm{C3}-\mathrm{output}
\]
requires on average 118 million operations to evaluate, and produced
a test error of 4.93\%. We deliberately kept the input spatial size
the same, so any improvements would be due to the introduction of
the time dimension.

\begin{figure}
\begin{centering}
\includegraphics[width=0.45\textwidth]{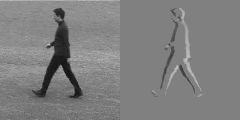}~~~\includegraphics[width=0.45\textwidth]{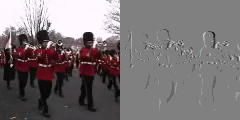}
\par\end{centering}

\caption{An image from the two video datasets used in Section \ref{sec:hra}, and the difference between that
frame and the previous frame.}
\end{figure}

\subsection{Human action recognition}\label{sec:hra}

Recognizing actions in videos is another example of a 2+1 dimensional
space-time problem. A simple way of turning a video into a sparse
3D object is to take the difference between successive frames, and
then setting to zero any values with absolute value below some threshold.
We tried this approach on two datasets, the simpler RHA dataset\footnote{\url{http://www.nada.kth.se/cvap/actions/}} \cite{CVAPDataset_RecognisingHumanAction} with 6 classes
of actions, and the harder UCF101 \cite{UCF101} dataset. We scaled the UCF101 video down by 50\% to have the same size as the HRA videos, $160\times 120$.
In both
cases we used a cubic CNN:
\[
32\mathrm{C}2-\mathrm{MP}3/2-64\mathrm{C}2-\mathrm{MP}3/2-\dots-192\mathrm{C2}-\mathrm{MP3/2-224C2-output.}
\]
For RHA, (mean) accuracy of 71.7\% is reported in \cite{CVAPDataset_RecognisingHumanAction}.
We used a threshold of 12\%, which resulted in 1.7\% of input pixels being active.
Our approach yielded 88.0\% accuracy with a computational cost of 1.1 billion
operations per test case.

For UCF101, accuracy of 43.90\% is reported
in \cite{UCF101}.
We used a threshold of 13\%, which resulted in 3.1\%  of input pixels being active.
The computational cost was higher than for RHA, 2.7 billion operatons, as the videos are more complicated.
Single testing produced an accuracy of 60.4\%, rising to 67.8\% with 12-fold testing.

These results are not state of the art. However, they do seem to strike a good balance in terms of computational cost. Also, we have not done any work to try to optimize our results. There are different ways of encoding a video's `optical flow' that we have not had a chance to explore yet.

\section{Conclusion}
We have shown that sparse 3D CNNs can be implemented efficiently, and produce interesting results for a variety of types of 3D data. There are potential applications that we have not yet tried. In biochemisty, there are large databases of 3D molecular structure. Proteins that are encoded differently may fold to produce similar shapes with similar functions. In robotics, it is natural to build 3D models by combining one or more 2D images with depth detector databases. Sparse 3D CNNs could be used to analyse these models.


\begin{thebibliography}{10}

\bibitem{multicolumndeep}
D.~Ciresan, U.~Meier, and J.~Schmidhuber.
\newblock \href{www.idsia.ch/~juergen/cvpr2012.pd}{Multi-column deep neural
  networks for image classification}.
\newblock In {\em Computer Vision and Pattern Recognition (CVPR), 2012 IEEE
  Conference on}, pages 3642--3649, 2012.

\bibitem{FractionalMaxPooling}
B.~Graham.
\newblock Fractional max-pooling, 2014.
\newblock
  \href{http://arxiv.org/abs/1412.6071}{http://arxiv.org/abs/1412.6071}.

\bibitem{GrahamSparse}
B.~Graham.
\newblock Spatially-sparse convolutional neural networks.
\newblock 2014.

\bibitem{DelvingDeeperRectifiers}
K.~He, X.~Zhang, S.~Ren, and J.~Sun.
\newblock Delving deep into rectifiers: Surpassing human-level performance on
  imagenet classification, 2014.
\newblock
  \href{http://arxiv.org/abs/1502.01852}{http://arxiv.org/abs/1502.01852}.

\bibitem{3dConvolutions}
S.~Ji, W.~Xu, M.~Yang, and K.~Yu.
\newblock 3d convolutional neural networks for human action recognition.
\newblock {\em IEEE Trans. Pattern Anal. Mach. Intell.}, 35(1):221--231, Jan.
  2013.

\bibitem{KarpathyCVPR14_LargeScaleVideoClassificationWithCNNs}
A.~Karpathy, G.~Toderici, S.~Shetty, T.~Leung, R.~Sukthankar, and L.~Fei-Fei.
\newblock Large-scale video classification with convolutional neural networks.
\newblock In {\em CVPR}, 2014.

\bibitem{UCF101}
A.~R.~Z. Khurram~Soomro and M.~Shah.
\newblock {U}{C}{F}101: A dataset of 101 human action classes from videos in
  the wild.
\newblock Technical report, November 2012.
\newblock CRCV-TR-12-01.

\bibitem{CIFAR10}
A.~Krizhevsky.
\newblock \href{http://www.cs.toronto.edu/~kriz/cifar.html}{Learning Multiple
  Layers of Features from Tiny Images}.
\newblock Technical report, 2009.

\bibitem{lecun-bengio-95a}
Y.~LeCun and Y.~Bengio.
\newblock Convolutional networks for images, speech, and time-series.
\newblock In M.~A. Arbib, editor, {\em The Handbook of Brain Theory and Neural
  Networks}. MIT Press, 1995.

\bibitem{CASIA}
C.-L. Liu, F.~Yin, D.-H. Wang, and Q.-F. Wang.
\newblock {CASIA} online and offline {C}hinese handwriting databases.
\newblock In {\em Proc. 11th International Conference on Document Analysis and
  Recognition (ICDAR), Beijing, China}, pages 37--41, 2011.

\bibitem{LecunFastTrainingFFTConvolutions}
M.~Mathieu, M.~Henaff, and Y.~LeCun.
\newblock Fast training of convolutional networks through ffts.
\newblock In {\em International Conference on Learning Representations
  (ICLR2014)}. CBLS, April 2014.

\bibitem{maturana_icra_2014}
D.~Maturana and S.~Scherer.
\newblock {3D Convolutional Neural Networks for Landing Zone Detection from
  LiDAR}.
\newblock In {\em {ICRA}}, 2015.

\bibitem{rigamonti2013learning}
R.~Rigamonti, A.~Sironi, V.~Lepetit, and P.~Fua.
\newblock Learning separable filters.
\newblock In {\em Computer Vision and Pattern Recognition (CVPR), 2013 IEEE
  Conference on}, 2013.

\bibitem{CVAPDataset_RecognisingHumanAction}
C.~Schuldt, I.~Laptev, and B.~Caputo.
\newblock Recognizing human actions: A {L}ocal {S}{V}{M} approach, 2004.

\bibitem{ShapeNet_Wu_2015_CVPR}
Z.~Wu, S.~Song, A.~Khosla, F.~Yu, L.~Zhang, X.~Tang, and J.~Xiao.
\newblock 3d shapenets: A deep representation for volumetric shapes.
\newblock In {\em The IEEE Conference on Computer Vision and Pattern
  Recognition (CVPR)}, June 2015.

\end{thebibliography}

\end{document}